\documentclass{article}
\usepackage{spconf}
\usepackage{amsmath}
\usepackage{amssymb}
\usepackage{subcaption}
\usepackage{diagbox}
\usepackage{url} 
\usepackage{graphicx} 

\usepackage{comment}
\usepackage{color}
\newcommand{\tana}[1]{\textcolor{black}{#1}} 

\usepackage{etoolbox}
\patchcmd{\thebibliography}
  {\settowidth}
  {\setlength{\itemsep}{0pt plus 0.3pt}\settowidth}
  {}
  {}

\newcommand\blfootnote[1]{
  \begingroup
  \renewcommand\thefootnote{}\footnote{#1}%
  \addtocounter{footnote}{-1}%
  \endgroup
}

\title{Adaptive Adversarial Cross-Entropy Loss \\ for Sharpness-Aware Minimization}
\name{Tanapat Ratchatorn and Masayuki Tanaka}
\address{Institute of Science Tokyo}

\begin{document}

\maketitle
\begin{abstract}
Recent advancements in learning algorithms have demonstrated that the sharpness of the loss surface is an effective measure for improving the generalization gap. Building upon this concept, Sharpness-Aware Minimization (SAM) was proposed to enhance model generalization and achieved state-of-the-art performance. SAM consists of two main steps, the weight perturbation step and the weight updating step. However, the perturbation in SAM is determined by only the gradient of the training loss, or cross-entropy loss. As the model approaches a stationary point, this gradient becomes small and oscillates, leading to inconsistent perturbation directions and also has a chance of diminishing the gradient. Our research introduces an innovative approach to further enhancing model generalization. We propose the Adaptive Adversarial Cross-Entropy (AACE) loss function to replace standard cross-entropy loss for SAM's perturbation. AACE loss and its gradient uniquely increase as the model nears convergence, ensuring consistent perturbation direction and addressing the gradient diminishing issue. Additionally, a novel perturbation-generating function utilizing AACE loss without normalization is proposed, enhancing the model's exploratory capabilities in near-optimum stages. Empirical testing confirms the effectiveness of AACE, with experiments demonstrating improved performance in image classification tasks using Wide ResNet and PyramidNet across various datasets. The reproduction code is available online~\footnote{\url{http://www.vip.sc.e.titech.ac.jp/proj/AACE}\\}.
\end{abstract}

\blfootnote{© 2024 IEEE. Personal use of this material is permitted. Permission from IEEE must be obtained for all other uses, in any current or future media, including reprinting/republishing this material for advertising or promotional purposes, creating new collective works, for resale or redistribution to servers or lists, or reuse of any copyrighted component of this work in other works.}

\section{Introduction}
\label{sec:intro}
In the recent development of machine learning, there has been a noticeable trend where models are becoming highly overparameterized. While these models are excellent at memorizing training data, a significant challenge arises in their performance on new, unseen data. This problem, known as overfitting, leads to a notable gap in performance between training and testing datasets \cite{zhang2021understanding}. Understanding how to improve the generalization of these models is crucial, as it can help them perform well not just on the data they were trained on, but also on new data they have never seen before.

To address the issue of generalization, researchers have explored various approaches. Some have taken a Bayesian perspective to understand this problem \cite{mcallester1999pac,neyshabur2017pac}, while others have looked at it through the information theory \cite{liang2019fisher}. 
Other significant areas of research are to investigate the impact of learning rate~\cite{li2019towards, DBLP:journals/corr/abs-1710-11029,DBLP:journals/corr/GoyalDGNWKTJH17} and batch size \cite{keskar2016large} on a model's generalization ability. Numerous techniques have been proposed to improve model generalization. Entropy-SGD uses local entropy \cite{chaudhari2019entropy}. Using Adam~\cite{kingma2014adam} as an optimizer in early training and switching to SGD~\cite{robbins1951stochastic} in later phases is also proven to improve generalization \cite{keskar2016large}. Integrating a partial adaptive parameter to the adaptive gradient methods such as Adam, Amsgrad was also introduced \cite{DBLP:journals/corr/abs-1806-06763}. Moreover, FOCA which avoids co-adaptation between a feature extractor and a particular classifier is another way to improve generalization \cite{DBLP:journals/corr/abs-1906-01150}.

\begin{figure}
    \centering
    \includegraphics[width=0.9\linewidth]{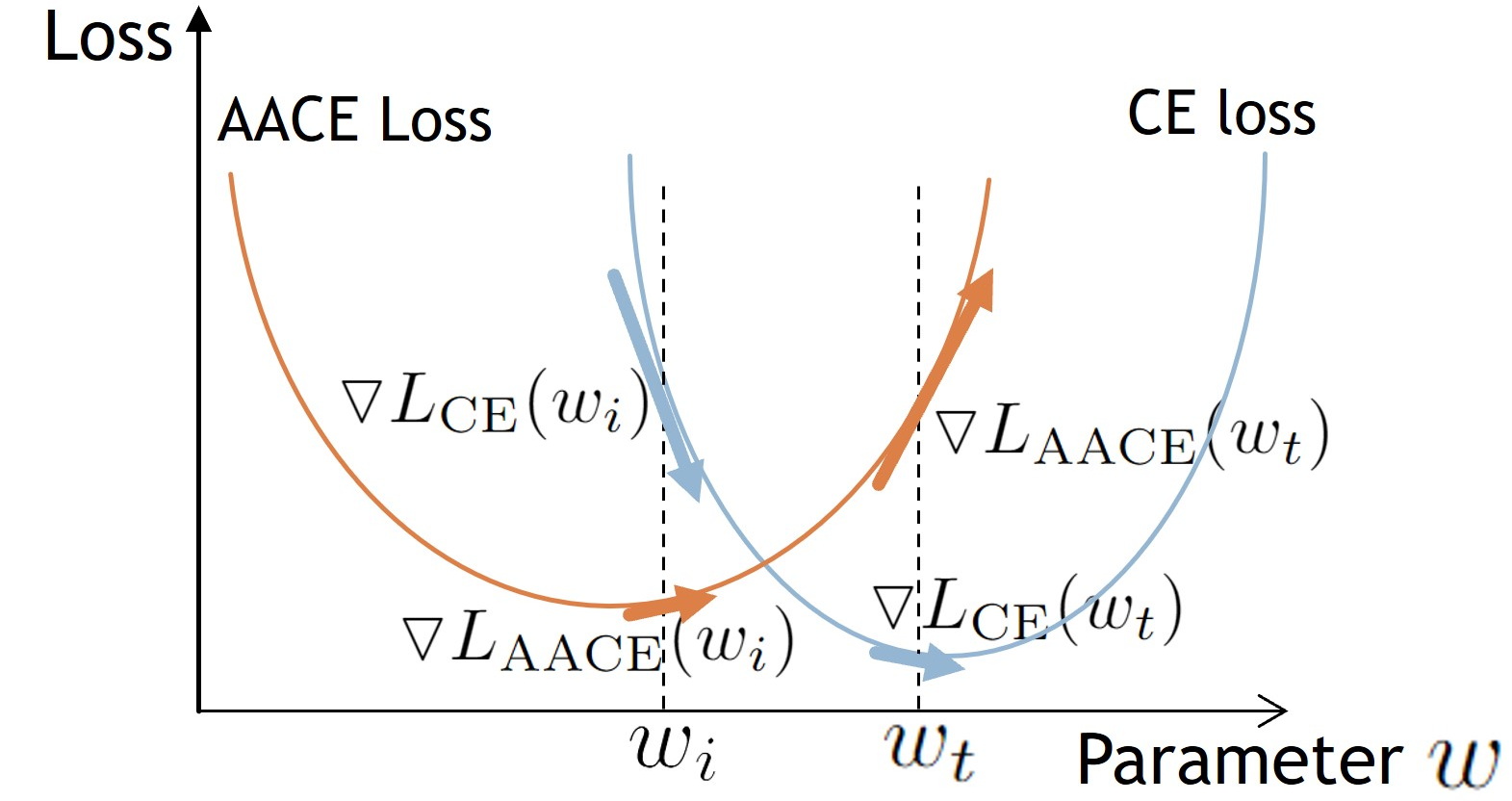}
    \caption{Comparison of loss and gradient between standard cross-entropy loss and Adaptive Adversarial Cross-Entropy loss at early stage ($w_{i}$) and later stage ($w_{t}$) of training.}
  
    \label{fig:1}
\end{figure}

Another important aspect of research focuses on the techniques related to the shape of the loss landscape and its connection to model generalization. Studies have shown that the sharpness of the loss surface and the minimization of derived generalization bounds are crucial for achieving superior performance in various tasks \cite{dziugaite2017computing, keskar2016large, hochreiter1997flat}. 

Developing efficient algorithms that aim for flatter minima, which in turn could improve generalization, remains a challenging area of research. Recently, Sharpness-Aware Minimization (SAM), an algorithm to search for flatter areas by adding small perturbations to the model parameters, was proposed and has proven to be generic and effective on several datasets and model architectures \cite{foret2020sharpness}.

SAM seeks a flat landscape by modifying the optimization process to explicitly consider the sharpness of the minima. SAM's algorithm can be decomposed into \tana{two} main steps. First, it finds a parameter configuration (weights) where the loss is high within a small neighborhood around the current weights. Then, it minimizes the model loss by using the gradient at this worst-case configuration.

Several novel methods also improve SAM's generalization performance further. GSAM introduced a sharpness measurement called surrogate gap \cite{zhuang2022surrogate}. PoF introduced a technique that updates the feature extractor to search for a flatter minima \cite{sato2022pof}. The adaptive sharpness which is scale-invariant was also introduced in ASAM \cite{DBLP:journals/corr/abs-2102-11600}. GA-SAM is another work that analyzes the relationship between local minima and generalization ability \cite{zhang2022ga}.

In this research, we found that while SAM has shown promising performance, there are still some issues to be concerned about. In finding the worst-case parameters, SAM's perturbation depends on the normalized gradient of cross-entropy loss and a pre-defined constant radius of the neighborhood. Since at the nearly optimum stage, the gradient of cross-entropy loss is very small and fluctuates around the optimum point, this leads to the unstable direction of the perturbation. Another noticeable issue is that, at the nearly optimum stage, the magnitude of the gradient of cross-entropy loss becomes smaller and smaller and has a risk of being zero which could cause devising by zero problem.

We propose a new approach to mitigate those issues by modifying loss in SAM's perturbation step. Instead of using cross-entropy loss that gets smaller as the model is trained, we introduce a new loss, Adaptive Adversarial Cross-Entropy (AACE) that grows as the model converges.

As demonstrated in Fig.~\ref{fig:1}, at the early stage of the training ($w_{i}$) both the loss and the magnitude of the gradient of the standard cross-entropy loss are high, and decrease as the model approaches convergence ($w_{t}$). On the contrary, AACE loss and its gradient magnitude start low and increase over the training process. This growing loss helps avoid the risk of the gradient diminishing at the saturated stage and leads to a more consistent direction of the gradient and the perturbation.

With this new loss, we also proposed not to normalize the loss in the perturbation step, making the perturbation not dependent on only a pre-defined constant. The new method also enlarges the magnitude of the perturbation step as the model converges, making the training more explorative even at the nearly optimum stage.

\section{Preliminary}
\label{sec:pre}
In traditional training of deep neural networks, optimization techniques like Stochastic Gradient Descent (SGD) seek to minimize the loss function. However, this process may converge to sharp minima, which are points in the parameter space where the loss is low for the training data but potentially high for unseen data. Sharp minima are believed to be less robust and generalize worse compared to flat minima.

Sharpness-Aware Minimization (SAM) is a novel training methodology designed to enhance the generalization \tana{performance} of deep learning models. Traditional training methods often converge to sharp minima, leading to suboptimal generalization. SAM, however, aims to find parameters that reside in neighborhoods having uniformly low loss, thus avoiding sharp minima. This is achieved through a min-max optimization problem efficiently solvable via gradient descent.

Instead of trying to minimize the loss as in vanilla training, SAM's objective is to minimize the perturbed loss which can be described as:
\begin{equation}\label{eq:1}
 L_{\rm SAM}(w) =  \max_{\left\|\varepsilon\right\|\leqslant\rho} L_{s}(w+\varepsilon)\,,
\end{equation}
where $L_{s}(w)$ is the training loss, $w$ represents the model parameters, and $\varepsilon$  is a perturbation vector bounded by $\rho$ in the L2-norm. The optimization seeks parameters $w$ such that the loss is minimized not just at $w$ but in its neighborhood within a radius of $\rho$.

In the case of small $\rho$, applying Taylor expansion around $w$, the $\varepsilon$ that satisfied the inner maximization in Eq.~\ref{eq:1} can be expressed as:

\begin{equation}\label{eq:2}
\varepsilon = {\rm StopGrad} \left( \rho\frac{\triangledown L_{s}(w)}{\left\|\triangledown L_{s}(w)\right\|_{2}} \right) \,,
\end{equation}
\tana{where ${\rm StopGrad}$ represents the stop graduation operation.} 
Note that ${\rm StopGrad}$ is not necessary to consider the inner maximization in Eq.~\ref{eq:1}. 
But we put ${\rm StopGrad}$ for the later discussion.
This formula determines the direction in the parameter space where the loss increases most sharply, scaled by the hyperparameter $\rho$. The ${\rm StopGrad}$ function is added here to ensure that this $\varepsilon$ is used only for the perturbation step and is treated as a fixed quantity during the computation of gradients for weight updates. 

\begin{figure*}
    \centering
    
    \begin{subfigure}[t]{0.33\textwidth}
        \centering
        \includegraphics[width=1\linewidth]{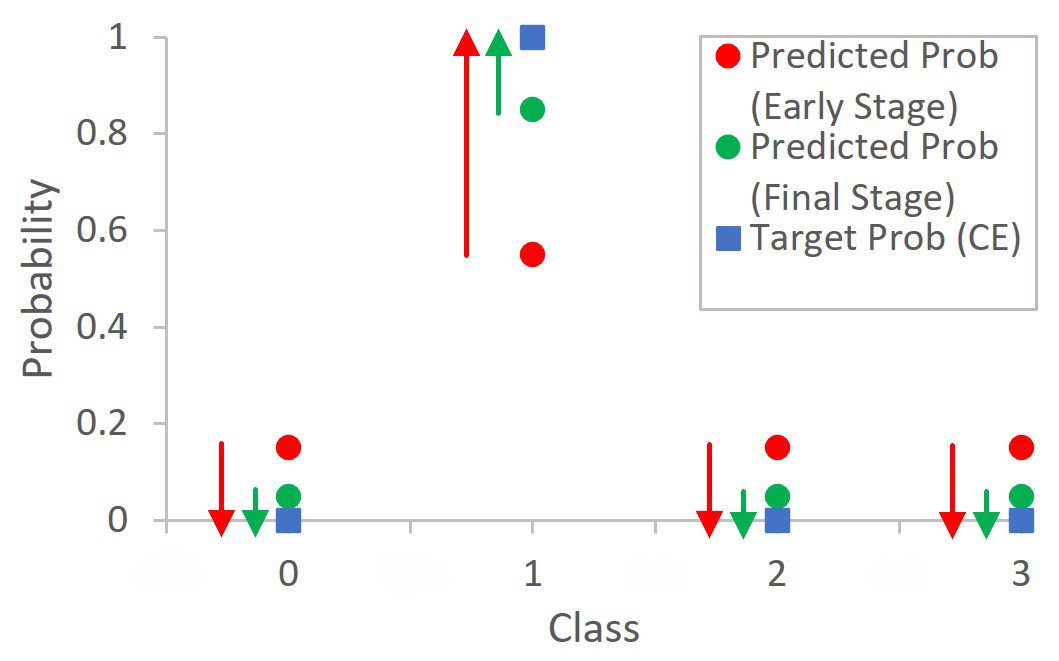}
        \caption{Probability distribution of standard cross-entropy loss.}
    \end{subfigure}%
    ~ 
    \begin{subfigure}[t]{0.33\textwidth}
        \centering
        \includegraphics[width=1\linewidth]{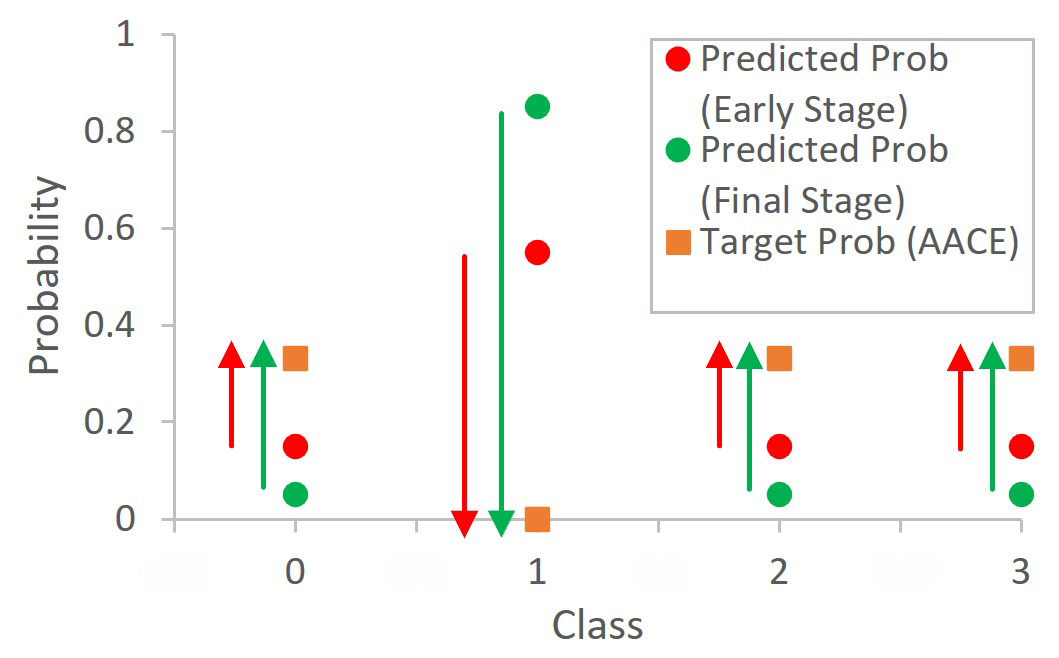}
        \caption{Probability distribution of Adaptive Adversarial Cross-Entropy loss.}
    \end{subfigure}
    ~ 
    \begin{subfigure}[t]{0.3\textwidth}
        \centering
        \includegraphics[width=1\linewidth]{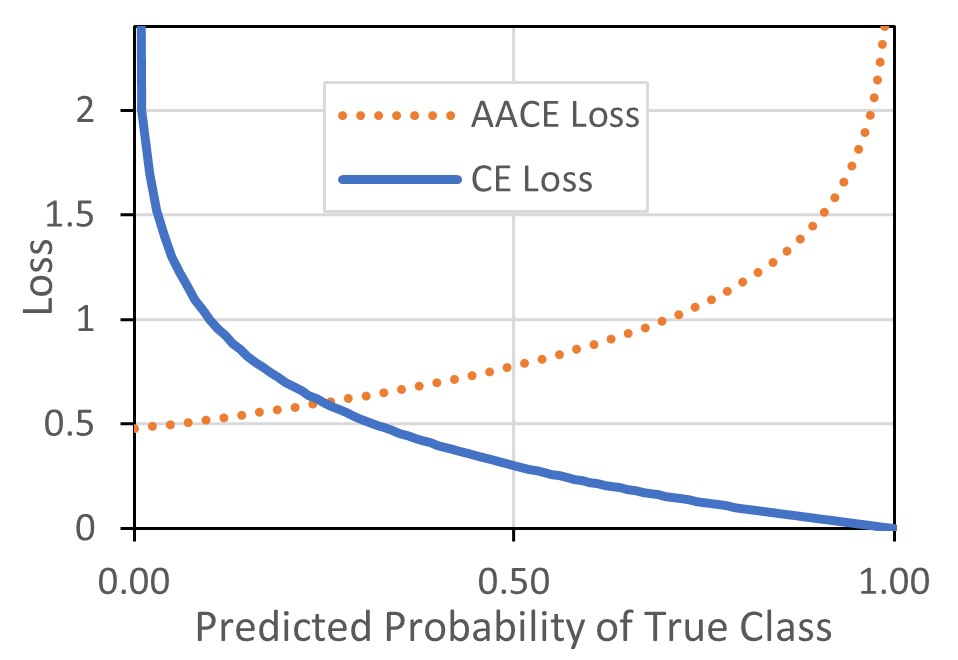}
        \caption{Loss trends of using standard cross-entropy loss vs ours.}
    \end{subfigure}
    
    \caption{Probability distributions and trend patterns of standard cross-entropy loss and Adaptive Adversarial Cross-Entropy loss.}
    \label{fig:2}
\end{figure*}

SAM's algorithm includes two main steps. First, the algorithm finds a worst-case perturbation of the current parameters using the perturbation vector calculated from Eq. \ref{eq:2}. Then, it updates the model weights by optimizing the model parameters using the gradients of the loss at the calculated perturbed position, as shown in the equation below.

\begin{equation}
w_{t+1} = w_{t}-\eta\triangledown L_{s}(w_{t}+ \varepsilon)\,,
\end{equation}
while $\eta$ is the learning rate. Note that for simplicity's sake, the weights updating formula is based on standard SGD without the momentum. 
However, in practical applications, alternative optimization algorithms such as Adam, RMSprop, or SGD with momentum can also be applied.

This approach encourages the optimizer to find flatter minima, which are believed to generalize better to unseen data. SAM has been shown to improve the performance of various deep learning models across different tasks, such as image classification.

\section{Proposed Method}
\label{sec:propose}
Although SAM demonstrates encouraging results, it's important to be aware of certain limitations. The method SAM uses to determine the worst-case parameters relies on a perturbation that is based on the normalized gradient of the cross-entropy loss, coupled with a predetermined constant defining the neighborhood's radius.

In this research, we consider Eq. \ref{eq:2} as a composition of $\rho$ and a specific function designed for constructing a perturbation vector.

\begin{equation}
\varepsilon = {\rm StopGrad}(\rho \, g(w))\,,
\end{equation}
\tana{where} $g(w)$ is named a perturbation generating function. 
In SAM, this function is described as
\begin{equation}
g_{\rm SAM}^{n}(w) = \frac{\triangledown L_{s}(w)}{\left\|\triangledown L_{s}(w)\right\|_{2}}\,,
\end{equation}
\tana{where we put the superscript $n$ because of the normalization.}

Consequently, for SAM, the perturbation direction is solely dependent on the gradient of the training loss $L_{s}(w)$, or the cross-entropy loss. At nearly stationary points, the gradient of cross-entropy loss becomes minuscule and oscillates around the optimal point, resulting in an inconsistent perturbation direction. Additionally, a significant concern arises when approaching stationary points. 
Here, the gradient of the cross-entropy loss tends to diminish, potentially reaching zero. This diminishing gradient poses a risk of a divide-by-zero error, which is a critical aspect to consider in the algorithm's application. Moreover, since the perturbation is always in a constant small radius, it is less explorative at the nearly optimum point.

On the other hand, we suggest the \tana{suitable} properties of the perturbation, especially, at the nearly stationary points. These properties include
\begin{enumerate}
  \item The direction of the perturbation should be sufficiently stable to meaningfully adjust the parameters.
  \item The gradient of loss used for perturbation calculation should not be too small and continuously decrease at the nearly optimum stage to avoid the gradient diminishing problem.
  \item The magnitude of the perturbation should be large enough to remain explorative while the model converges.  
\end{enumerate}

Hence, we introduce an innovative method to address the challenges associated with SAM's perturbation step and satisfy the required properties of the perturbation. Our approach involves altering the loss function used for calculating the perturbation vector. Rather than relying on the cross-entropy loss, which diminishes as the model trained, we propose a novel loss function named Adaptive Adversarial Cross-Entropy (AACE). This new loss function is designed to increase magnitude as the model approaches convergence.

According to the calculation of cross-entropy loss which is determined as

\begin{equation}
L = -\sum_{i} \tau_{i} \log(q_i)\,,
\end{equation}
where $q_{i}$ is the predicted probability corresponding to class $i$, and $\tau_{i}$ is the target probability distribution for class $i$, which, for the standard cross-entropy loss, can be determined using one-hot encoding as

\begin{equation}
\tau_{i}^{\rm CE} =
  \begin{cases}
              1, &         (i=t)\\
              0, &         (i\neq t)\\
      \end{cases}\,,
\end{equation}
where $t$ stands for a \tana{ground truth} class.

In our proposed method, instead of using hard \tana{0} or \tana{1} as a target for \tana{ground truth} and negative classes as in the standard cross-entropy, Adaptive Adversarial Cross-Entropy (AACE) defines new adversarial labels. For a positive class, the label is set to 0. On the other hand, for negative classes, a new target distribution is adjusted by the ratio of the predicted probability $q_{i}$ of a specific class $i$ to the sum of predicted probabilities of all negative classes.

\begin{equation}
\tau_{i}^{\rm AACE} = \xi(\tilde{q_{i}})\,,
\end{equation}
given
\begin{equation}
\xi(\tilde{q_{i}}) =
  \begin{cases}
              0, &  (i = t)\\
              \frac{\tilde{q_{i}}}{\sum_{i \neq t}\tilde{q_{i}}}, & (i \neq t)\\
      \end{cases}\,,
\end{equation}
and
\begin{equation}
\tilde{q_{i}} = {\rm StopGrad}(q_{i})\,.
\end{equation}

As a result, our proposed adaptive adversarial labeling keeps the calculated loss high thanks to enlarging the difference between the target probability distributions and the model's predicted probability distributions. As illustrated in Fig.~\ref{fig:2}, assuming that the predicted probabilities are equally distributed among the negative classes, in standard cross-entropy loss with one-hot encoding targets (Fig.~\ref{fig:2}~(a)), the differences between the predicted probabilities and the target probabilities decrease as the model converges. In contrast, with AACE, these differences increase as the model converges (Fig.~\ref{fig:2}~(b)). Also, while the standard cross-entropy loss decreases as the predicted probability of the positive class approaches 1, the AACE loss, conversely, increases as the predicted probability of the positive class nears 1 (Fig.~\ref{fig:2}~(c)).

\begin{figure}
    \centering
    \includegraphics[height=4.8cm]{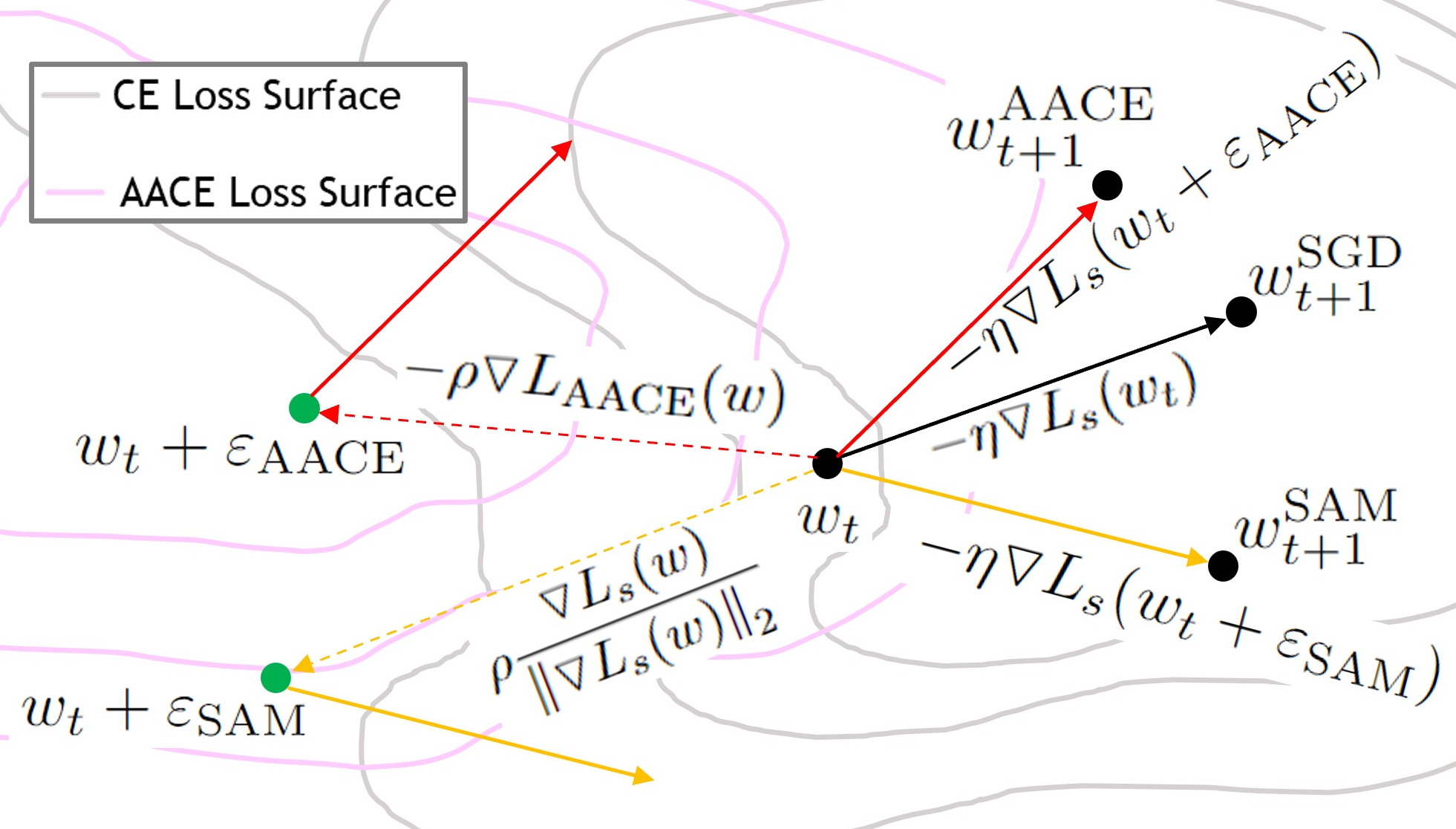}
    \caption{Diagram illustrates the perturbation step and the updating step of original SAM and our proposed method.}
    \label{fig:3}
\end{figure}

Moreover, it is well-known that the gradient of cross-entropy loss with respect to the logit before the softmax activation can be calculated from:

\begin{equation}
\frac{\partial L}{\partial z_i} = q_i - \tau_i\,.
\end{equation}
In which $\frac{\partial L}{\partial z_i}$ represents the gradient of the loss with respect to the logits $z_i$ for class $i$. $q_i$ is the predicted probability for class $i$, as outputted by the softmax function applied to the logits.

While the gradient of conventional cross-entropy loss decreases as the model converges, our AACE loss with adversarial targets increases due to the growth of the gaps between the predicted probabilities and the newly defined adversarial labels. Hence, the gradient for AACE loss remains high even at a nearly optimum stage. 

As a result of the increase in the perturbation loss and its gradient while the model converges, the risk of gradient diminishes, which leads to devising by zero problem, is eliminated. More importantly, the larger and growing gradient gives rise to a stronger and more stable direction of the perturbation in SAM's perturbation step.

In order to define our perturbation generating function, given that

\begin{equation}
L_{\rm AACE}(w) = -\sum_{i} \tau_{i}^{\rm AACE} \log(q_i)\,,
\end{equation}
the perturbation generating function can now be defined as

\begin{equation}
g_{\rm AACE}^{n}(w) = -\frac{\triangledown L_{\rm AACE}(w)}{\left\|\triangledown L_{\rm AACE}(w)\right\|_{2}}\,.
\end{equation}
Now that our trends of perturbation loss and gradient are converse to the original SAM, the negative sign is applied here because we need to perturb the model parameters to the worst configuration in which AACE loss is low, as opposed to the case of using normal cross-entropy loss.

Furthermore, since we prefer to enlarge the magnitude of the perturbation as the model converges, we also proposed to not normalize the gradient and define a new perturbation generating function as

\begin{equation}
g_{\rm AACE}(w) = -\triangledown L_{\rm AACE}(w)\,.
\end{equation}

Due to the nature of AACE loss, this newly defined perturbation vector guarantees to increase and remain consistence in direction, even at the nearly optimum stage.

Finally, the weights updating of SGD, original SAM, and our SAM with AACE can be represented by the following expressions. 

\begin{align}
w_{t+1}^{\rm SGD} &= w_{t}-\eta\triangledown L_{s}(w_{t})\,, \\
w_{t+1}^{\rm SAM} &= w_{t}-\eta\triangledown L_{s}(w_{t}+\varepsilon_{\rm SAM})\,, \\
w_{t+1}^{\rm AACE} &= w_{t}-\eta\triangledown L_{s}(w_{t}+\varepsilon_{\rm AACE})\,,
\end{align}
where
\begin{equation}
\varepsilon_{\rm SAM} = {\rm StopGrad} \left(\rho \frac{\triangledown L_{s}(w)}{\left\|\triangledown L_{s}(w)\right\|_{2}} \right)\,,
\end{equation}
and
\begin{equation}
\varepsilon_{\rm AACE} = - {\rm StopGrad}(\rho \triangledown L_{\rm AACE}(w))\,.
\end{equation}

The comparison of these weight updates is shown in Fig.~\ref{fig:3}.  Instead of updating the model configuration based on the loss's gradient at the current position as in SGD, SAM and our proposed method slightly perturb the model weights to a new position. Then the gradient at the perturbed weights is calculated. This calculated gradient is used to update the model weight at the current configuration.

\begin{table}[t!]
\caption{Accuracies ($\%$) of SAM with AACE on Wide ResNet on CIFAR-100 with different $\rho$, with and without gradient normalization in the perturbation.\\}
\vspace*{-4mm}
\centering
\begin{tabular}{ |c|c c| } 
 \hline
  \diagbox{$\rho$}{$g(w)$} & $-\frac{\triangledown L_{\rm AACE}(w)}{\left\|\triangledown L_{\rm AACE}(w)\right\|_{2}}$ & $-\triangledown L_{\rm AACE}(w)$ \\ 
 \hline
 0.05 & 82.11 & 83.82 \\ 
 0.1 & 83.19 & 84.09 \\ 
 0.2 & 83.66 & \textbf{84.33} \\ 
 0.5 & 84.13 & 84.02 \\
 1.0 & 84.10 & 84.23 \\
 2.0 & 70.08 & 78.56 \\
 5.0 & 27.38 & 71.19 \\
 \hline
\end{tabular}
\label{table:1}
\end{table}

\section{Experiments}
\label{sec:experiments}
In order to prove AACE performance, empirical research has been conducted on several model architectures and datasets.

\subsection{Hyperparameter grid search}
\label{ssec: grid search}
First of all, to evaluate the performance of SAM with AACE loss, we trained Wide ResNet \cite{zagoruyko2016wide} on the CIFAR-100 \cite{krizhevsky2009learning} dataset. We used model depth = 28, width factor = 10, and used SGD as a base optimizer. We applied horizontal flip, padding by four pixels, and random crop for data augmentations. Cutout regularization \cite{devries2017improved} was also applied. SAM's only hyperparameter, $\rho$, 
has been tuned via grid search over \{0.05, 0.1, 0.2, 0.5, 1.0, 2.0, 5.0\}. We trained the models for 200 epochs with batch size = 256, momentum = 0.9, and weight decay = 0.0005. We set the initial learning rate to 0.1 and drop by 0.2 at 30\%, 60\%, and 80\% of the training. The experiments were conducted using both perturbations with and without gradient normalization.

As seen in Table 1, the experiments in which we did not apply gradient normalization tend to have higher accuracy. This aligns with our hypothesis. Our proposed AACE elevates the gradient during training. When the gradient normalization part is applied, the perturbation's magnitude consistently remains the same. Conversely, our proposed method suggests to remove the gradient normalization from the perturbation which leads to an increase in its magnitude. Hence, this approach makes the model to be more explorative at the nearly optimum stage.

Furthermore, during this grid search, the experiment that uses $\rho = 0.2$ without gradient normalization shows the best performance. We hence use this parameter for the following experiments.

\begin{figure}
    \centering
    
    \begin{subfigure}[t]{0.235\textwidth}
        \centering
        \includegraphics[width=1\linewidth]{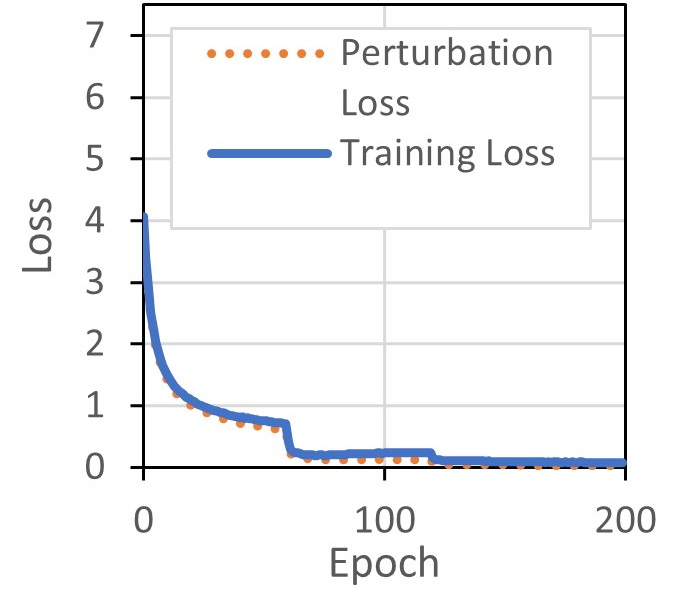}
        \caption{Perturbation loss and training loss of original SAM.}
    \end{subfigure}%
    ~ 
    \begin{subfigure}[t]{0.235\textwidth}
        \centering
        \includegraphics[width=1\linewidth]{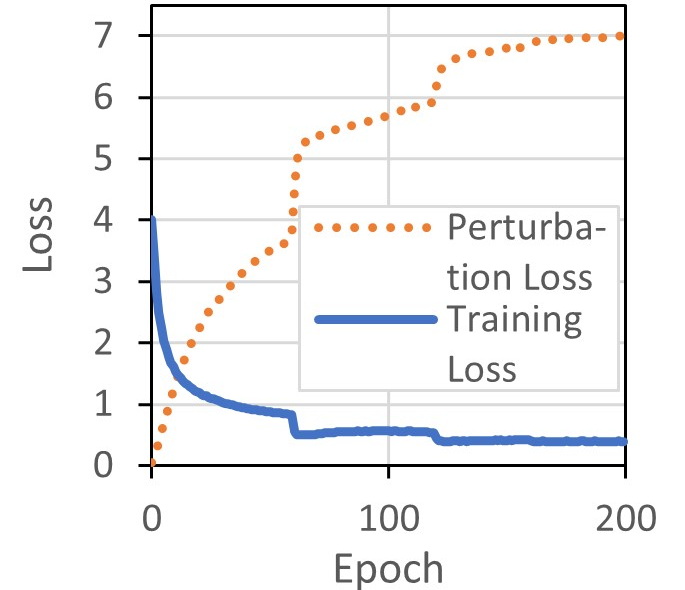}
        \caption{Perturbation loss and training loss of SAM with AACE.}
    \end{subfigure}
    
    \caption{Losses comparison of standard SAM and SAM with AACE. Each data point is the average
loss in the epoch.}
    \label{fig:4}
\end{figure}

\subsection{Effect of using AACE for perturbation loss}
\label{ssec:effect_of_aace}
To prove the properties of AACE as discussed in the previous section, the experiments were conducted on the Wide ResNet on the CIFAR-100 dataset using the original SAM and our proposed method. We used the same values for all hyperparameters. For original SAM $\rho$ was set to 0.05, the same as in SAM's original paper. For our proposed method, we set $\rho$ to 0.2.

Firstly, we investigated the characteristics of standard cross-entropy loss and our adaptive adversarial cross-entropy loss. In standard SAM (Fig.~\ref{fig:4}~(a)), the curve of the average perturbation loss aligns with the trend of training loss. However, when utilizing SAM with AACE (Fig.~\ref{fig:4}~(b)), the average perturbation loss increases as the model converges, which is against the trend of the training loss. Note that the rapid rises/drops of the curves are caused by the learning rate scheduler.

\begin{figure}
    \centering

    \begin{subfigure}[t]{0.235\textwidth}
        \centering
        \includegraphics[width=1\linewidth]{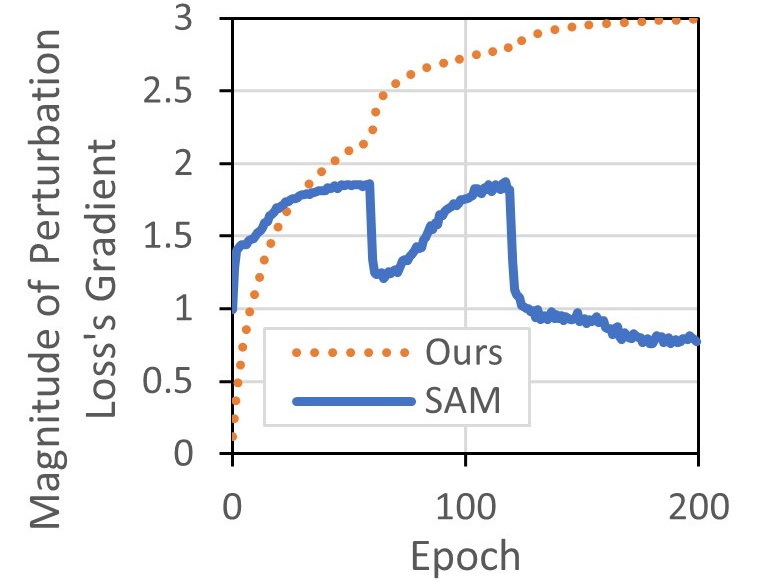}
        \caption{Magnitudes of perturbation loss's gradients of SAM and the proposed method.}
    \end{subfigure}\hfill%
    ~ 
    \begin{subfigure}[t]{0.215\textwidth}
        \centering
        \includegraphics[width=1\linewidth]{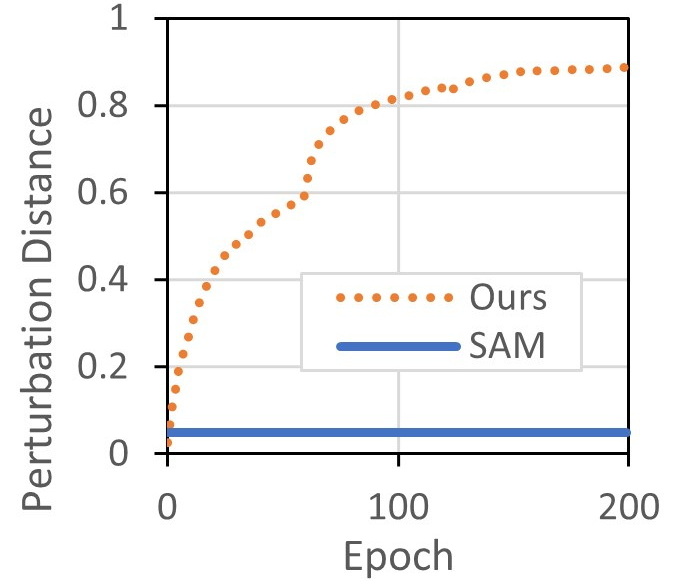}
        \caption{Perturbation distances of SAM and the proposed method.}
    \end{subfigure}
    
    \caption{Comparison of magnitudes of perturbation loss’s gradients and perturbation distances between SAM and our method. Note that each data point represents the average value of samples in the epoch.}
    \label{fig:5}
\end{figure}

Moreover, the average magnitudes of the gradient of perturbation loss are compared as shown in Fig.~\ref{fig:5}~(a). In the standard SAM, the gradient magnitude of the perturbation loss tends to decrease as the model nears convergence. However, when SAM is integrated with AACE, this gradient magnitude shows an increase as the model approaches convergence. Since the magnitude of the gradient of the perturbation remains high, it leads to a more stable gradient direction and also avoids the gradient diminishing issue. Also, Fig.~\ref{fig:5}~(b) shows that the average perturbation distances are consistently equal to $\rho$ (0.05) in the original SAM, due to gradient normalization. However, our approach suggests not normalizing the perturbation loss's gradient, leading to an increase in the average perturbation distances as the model progresses toward convergence thanks to the increase in magnitudes of the gradients of perturbation loss. We believe that this could lead the model to be more explorative at the final stage of training.

\begin{figure}
    \centering
    \includegraphics[width=0.9\linewidth]{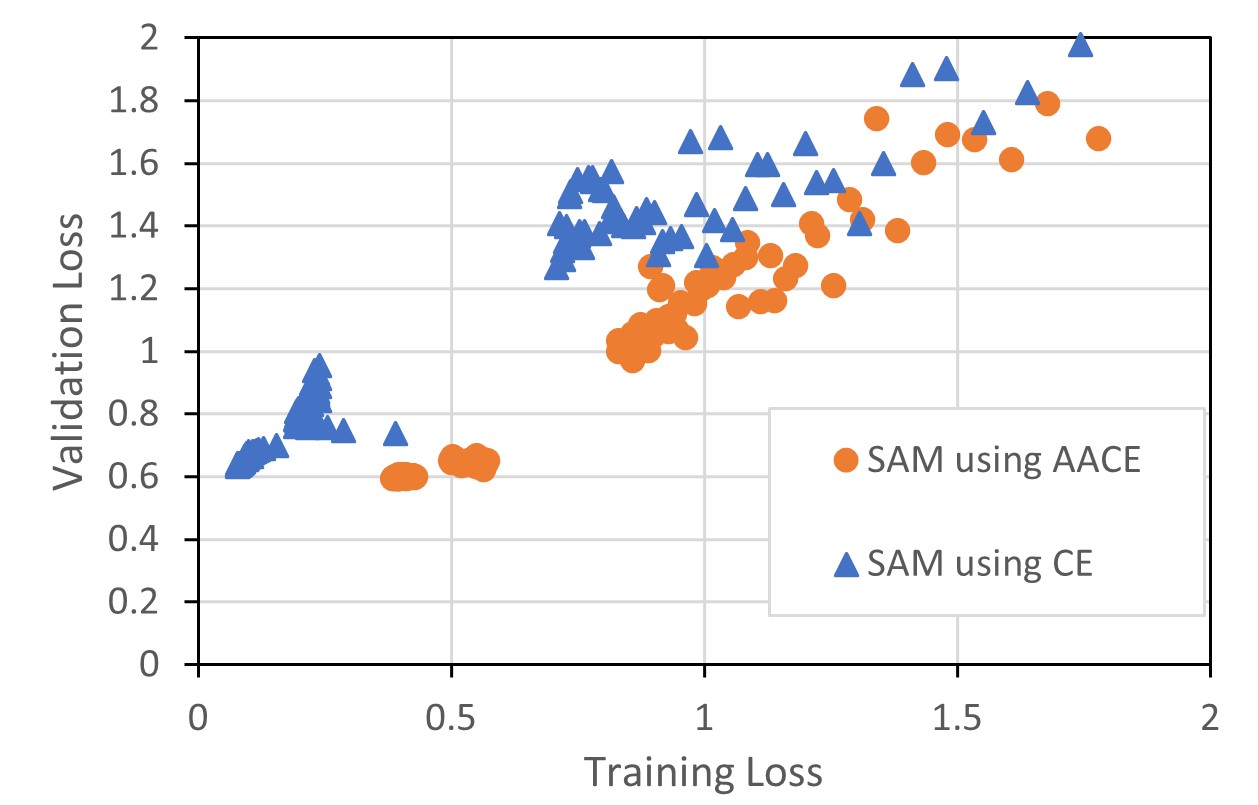}
    \caption{Validation loss and training loss comparison between the models trained with SAM using CE loss and AACE loss in perturbation step. Each data point represents the average training/validation loss in the epoch.}
    \label{fig:6}
\end{figure}

More importantly, as shown in Fig.~\ref{fig:6}, we explored the generalization ability of the original SAM versus our method. When comparing the performance of the models trained with SAM using standard cross-entropy (CE) loss and Adaptive Adversarial Cross-Entropy (AACE) loss for the perturbation step, it is noticed that while the model trained on SAM with CE achieve a lower training loss, SAM with AACE shows a lower validation loss. This indicates that SAM integrated with AACE loss exhibits superior generalization capabilities.

\subsection{Image Classification \tana{Comparisons} with Wide ResNet}
\label{ssec:WRN}

To confirm the effectiveness of SAM integrated with AACE, we conducted empirical experiments on Wide ResNet with different datasets such as CIFAR-100, CIFAR-10 \cite{krizhevsky2009learning}, Fashion-MNIST \cite{xiao2017fashion}, and Food101 \cite{bossard14}. All the hyperparameters are the same as in the previous experiment. The models were trained for 200 epochs for the original SAM and proposed method and 400 epochs for vanilla SGD since SAM weight update requires twice backpropagation compared to SGD. The results for SGD, SAM, and SAM with AACE are shown in Table 2. As seen in the table, our proposed method beat SGD and original SAM on all datasets.

\subsection{Image Classification \tana{Comparisons} with PyramidNet}
\label{ssec:PRNl}

We also observed the performance of our proposed methods on different model architecture, PyramidNet \cite{han2017deep}. In this experiment, the previous datasets were used. For the model setup, we used PyramidNet network with depth = 272, alpha = 200, and batch size = 64. The rest hyperparameters, including the number of epochs, are the same as in the previous experiment. Similar to the experiment on Wide ResNet, as seen in Table 3, our proposed method revealed the highest accuracies for most datasets, except for the CIFAR-10.

\begin{table}[t!]
\caption{Accuracies ($\%$) of models training with SGD, SAM, and our proposed method on Wide ResNet\\}
\vspace*{-4mm}
\centering
\begin{tabular}{ |c|c c c| } 
 \hline
  Dataset & SGD & Original SAM & Proposed \\ 
 \hline
 CIFAR-100 & 82.21 & 83.52 &  \textbf{84.33} \\ 
 CIFAR-10 & 96.63 & 97.02 &  \textbf{97.04}\\ 
 Fashion-MNIST & 94.57 & 95.26 & \textbf{95.41}\\ 
 Food101 & 65.12 & 70.34 &  \textbf{73.55}\\
 \hline
\end{tabular}
\label{table:2}
\end{table}

\begin{table}[t!]
\caption{Accuracies ($\%$) of models training with SGD, SAM, and our proposed method on PyramidNet\\}
\vspace*{-4mm}
\centering
\begin{tabular}{ |c|c c c| } 
 \hline
  Dataset & SGD & Original SAM & Proposed \\ 
 \hline
 CIFAR-100 & 81.25 & 83.85 &  \textbf{84.13} \\ 
 CIFAR-10 & 95.74 & \textbf{96.95} & 96.52\\ 
 Fashion-MNIST & 95.03 & 95.51 & \textbf{95.57}\\ 
 Food101 & 66.43 & 72.97 &  \textbf{75.94}\\
 \hline
\end{tabular}
\label{table:3}
\end{table}

\section{Conclusion}
\label{sec:conclusion}

In conclusion, this research addresses the key limitations of Sharpness-Aware Minimization (SAM) and makes an improvement by proposing a novel perturbation generating technique. We introduce the Adaptive Adversarial Cross-Entropy (AACE) loss which can replace the standard cross-entropy loss in SAM's perturbation step. AACE loss and its gradient increase as the model approaches convergence, hence it ensures a more consistent direction of the perturbation and also prevents a gradient diminishing problem. We also suggest a new perturbation generating function that uses AACE loss without the normalization part, which increases the magnitude of the perturbation, making the model more explorative at the nearly optimum stage. The empirical results confirmed our hypothesis on AACE characteristics and the experiment results show that our proposed method helps SAM to perform better for image classification tasks on Wide ResNet and PyramidNet on various datasets.

\section{Acknowledgements}
\label{sec:acknowledgements}

This work was partially supported by Tateishi Research Grant (A) 2241011 and JSPS KAKENHI Grant Numbers 24K02957.

\bibliographystyle{IEEEbib}
\bibliography{ref}

\end{document}